\documentclass[11pt,a4paper]{article}
\usepackage[hyperref]{emnlp-ijcnlp-2019}
\usepackage{times}
\usepackage{latexsym}
\usepackage{graphicx}
\usepackage{url}
\usepackage{romannum}
\usepackage{amssymb}
\usepackage{hhline}
\usepackage{tabularx}
    \newcolumntype{L}{>{\raggedright\arraybackslash}X}
\usepackage{multirow, makecell}
\usepackage{rotating}
\usepackage{algorithm}
\usepackage{algpseudocode}
\aclfinalcopy 

\newcommand{\squeezeup}

\title{Extractive Summarization of Long Documents by Combining Global and Local Context}

\author{Wen Xiao and Giuseppe Carenini\\
  Department of Computer Science \\
  University of British Columbia \\
  Vancouver, BC, Canada, V6T 1Z4 \\
  {\tt \{xiaowen3, carenini\}@cs.ubc.ca}}

\date{}

\begin{document}
\maketitle
\begin{abstract}
In this paper, we propose a novel neural single-document extractive summarization model for long documents, incorporating both the global context of the whole document and the local context within the current topic. We evaluate the model on two datasets of scientific papers , Pubmed and arXiv, where it outperforms previous work, both extractive and abstractive models, on ROUGE-1, ROUGE-2 and METEOR scores. We also show that, consistently with our goal, the benefits of our method become stronger as we apply it to longer documents.
Rather surprisingly, an ablation study indicates that the benefits of our model seem to come exclusively from modeling the local context, even for the longest documents.
\end{abstract}

\section{Introduction}
\renewcommand{\thefootnote}{\Roman{footnote}}
Single-document summarization is the task of generating a short summary for a given document. Ideally, the generated summaries should be fluent and coherent, and should faithfully maintain the most important information in the source document. \textcolor{purple}{This is a very challenging task, because it arguably requires an in-depth understanding of the source document, and current automatic solutions are still far from human performance} \cite{survey}.\footnote[7]{Sentence coloring and Roman
numbering will be explained in the result sub-section 4.5.}


Single-document summarization can be either extractive or abstractive. Extractive methods typically pick sentences directly from the original document based on their importance, and form the summary as an aggregate of these sentences. Usually, summaries generated in this way have a better performance on fluency and grammar, but they may contain much redundancy and lack in coherence across sentences. In contrast, abstractive methods attempt to mimic what humans do by first extracting content from the source document and then produce new sentences that aggregate and organize the extracted information. Since the sentences are generated from scratch they tend to have a relatively worse performance on fluency and grammar. Furthermore, while abstractive summaries are typically less redundant,  they may end up including misleading or even utterly false statements, because the methods to extract and aggregate information form the source document are still rather noisy. 





In this work, we focus on extracting informative sentences from a given document (without dealing with redundancy), especially when the document is relatively long (e.g., scientific articles).

Most recent works on neural extractive summarization have been rather successful in generating summaries of short news documents (around 650 words/document) \cite{RNN_beyond} by applying neural Seq2Seq models \cite{cheng&lapata}. However when it comes to long documents, these models tend to struggle with longer sequences because at each decoding step, the decoder needs to learn to construct a context vector capturing relevant information from all the tokens in the source sequence \cite{shao}. 

Long documents typically cover multiple topics. In general, the longer  a document is, the more topics are discussed. As a matter of fact, when humans write long documents they organize them in chapters, sections etc.. Scientific papers are an example of longer documents and they follow a standard discourse structure describing the problem, methodology, experiments/results, and finally conclusions \cite{scientific_paper}.


To the best of our knowledge only one previous work in extractive summarization has explicitly leveraged section information to guide the generation of summaries \cite{section_supervised}. However, the only information about sections fed into
their sentence classifier is a categorical feature with values like \textit{Highlight}, \textit{Abstract},
\textit{Introduction}, etc.,
depending on which section the sentence appears in.

In contrast, in order to exploit section information, in this paper we propose to capture a distributed representation of both the global (the whole document) and the local context (e.g., the section/topic) when deciding if a sentence should be included in the summary


Our main contributions are as follows:
{\bf (i)} In order to capture the local context, we are the first to apply LSTM-minus to text summarization. LSTM-minus is a method for learning embeddings of text spans, which
has achieved good performance in dependency parsing
\cite{lstm-minus_propose},  in constituency
parsing \cite{lstm-minus_constituency},
as well as in discourse parsing \cite{lstm-minus_discourse}. With respect to more traditional methods for capturing
local context, which rely on hierarchical
structures, LSTM-minus produces simpler
models i.e. with less parameters, and therefore faster to train and less prone to overfitting.
{\bf (ii)} We test our method on the Pubmed and arXiv datasets and results appear to support our goal of effectively summarizing long documents. In particular, while overall we outperform the baseline and previous approaches only by a narrow margin on both datasets, the benefit of our method become much stronger as we apply it to longer documents. \textcolor{purple}{Furthermore, in an ablation study to assess the relative contributions of the global and the local model we found that, rather surprisingly, the benefits of our model seem to come exclusively from modeling the local context, even for the longest documents.}\footnotemark[6]
\renewcommand{\thefootnote}{\arabic{footnote}}
{\bf (iii)} In order to evaluate our approach, we have created oracle labels for both Pubmed and arXiv \cite{discourse_long_document}, by applying a greedy oracle labeling algorithm. The two datasets annotated with extractive labels will be made public.\footnote{The data and code are available at \url{https://github.com/Wendy-Xiao/Extsumm_local_global_context}.}

\section{Related work}
\subsection{Extractive summarization}
Traditional extractive summarization methods are mostly based on explicit surface features \cite{feature_based}, relying on graph-based methods \cite{textrank}, or on submodular maximization \cite{tixier17}. Benefiting from the success of neural sequence models in other NLP tasks, \newcite{cheng&lapata} propose a novel approach to extractive summarization based on neural networks and continuous sentence features, which outperforms traditional methods on the DailyMail dataset. In particular, they develop a general encoder-decoder architecture, where a CNN is used as sentence encoder, a uni-directional LSTM as document encoder, with another uni-directional LSTM as decoder. To decrease the number of parameters while maintaining the accuracy, \newcite{summarunner} present SummaRuNNer, a simple RNN-based sequence classifier without decoder, outperforming or matching the model of \cite{cheng&lapata}. They take content, salience, novelty, and position of each sentence into consideration when deciding if a sentence should be included in the extractive summary. Yet, they do not capture any aspect of the topical structure, as we do in this paper. So their approach would arguably suffer when  applied to long documents, likely containing multiple and diverse topics.

While SummaRuNNer was tested only on news,  \newcite{EMNLP2018} carry out  a comprehensive set of experiments with deep learning models of extractive summarization across different domains, i.e. news, personal stories, meetings, and medical articles, as well as across different neural architectures, in order to better understand the general pros and cons of different design choices. They find that non auto-regressive sentence extraction performs as well or better than auto-regressive extraction in all domains, where by auto-regressive sentence extraction they mean using previous predictions to inform future predictions. Furthermore, they find that the Average Word Embedding sentence encoder works at least as well as encoders based on CNN and RNN. In light of these findings, our model is  not auto-regressive and uses the Average Word Embedding encoder.
\subsection{Extractive summarization on Scientific papers}
Research on summarizing scientific articles has a long history \cite{survey_summarization}. Earlier on, it was realized   that summarizing scientific papers requires different approaches than what was used for summarizing news articles, due to differences in document length, writing style and rhetorical structure. For instance, \cite{scientific_article_summarization_02} presented a supervised Naive Bayes classifier to select content from a scientific paper based on the rhetorical status of each sentence (e.g., whether it specified a research goal, or some generally accepted scientific background knowledge, etc.). 
More recently, researchers have extended this work by applying more sophisticated classifiers to identify more fine-grain rhetorical categories, as well as by exploiting citation contexts. \newcite{2013-discourse} propose the CoreSC discourse-driven content, which relies on CRFs and SVMs, to classify the discourse categories (e.g. Background, Hypothesis, Motivation, etc.) at the sentence level.  The recent work most similar to ours is \cite{section_supervised} where, in order to determine whether a sentence should be included in the summary, they directly use the section  each sentence appears in as a categorical feature with  values like Highlight, Abstract, Introduction, etc.. In this paper, instead of using sections as categorical features, we rely on a distributed representation of the semantic information within each section, as the local context of each sentence. In a very different line of work, \newcite{cohan-2015-scientific} form the summary  by also exploiting information on how the target paper is cited in  other papers. Currently, we do not use any information from citation contexts.



\subsection{Datasets for long documents}
\begin{table}[t!]
    \centering
    \resizebox{\linewidth}{!}{
    \begin{tabular}{|c|c|c|c|}
    \hline
    Datasets & \# docs & avg. doc. length&  avg. summ. length\\
    \hline
      CNN &92K& 656& 43\\
      \hline
        Daily Mail & 219K & 693 & 52\\
        \hline
        NY Times & 655K & 530 & 38\\
        \hline
        PubMed & 133K & 3016& 203\\
        \hline
        arXiv & 215K & 4938 & 220 \\
        \hline
     \end{tabular}}
    \caption{Comparison of news datasets and scientific paper datasets\cite{discourse_long_document}, the length is in terms of the number of words}
    \label{tab:dataset}
\end{table}
\newcite{summary_dataset} provide a comprehensive overview of the current  datasets for summarization. Noticeably, most of the larger-scale summarization datasets consists of relatively short documents, like CNN/DailyMail \cite{RNN_beyond} and New York Times \cite{nyt}. One exception is \cite{discourse_long_document} that recently introduce two large-scale datasets of long and structured scientific papers obtained from arXiv and PubMed. These two new datasets contain much longer documents than all the news datasets (See Table \ref{tab:dataset}) and are therefore ideal test-beds for the method we present in this paper.

\subsection{Neural Abstractive summarization on long documents}
While most current neural abstractive summarization models have focused on summarizing relatively short news articles (e.g., \cite{get_to_the_point}), few researchers have started to investigate the summarization of longer documents by exploiting their natural structure. 
\newcite{agents} present an encoder-decoder architecture to address the challenges of representing a long document for abstractive summarization. The encoding task is divided across several collaborating agents, each is responsible for a subsection of text through a multi-layer LSTM with word attention.
Their model seems however overly complicated when it comes to the extractive summarization task, where word attention is arguably much less critical. So, we do not consider this model further in this paper. 

\newcite{discourse_long_document} also propose a model for abstractive summarization taking the structure of documents into consideration with a hierarchical approach, and test it on longer documents with section information, i.e. scientific papers. In particular, they apply a hierarchical encoder at the word and section levels. Then, in the decoding step, they combine the word attention and section attention to obtain a context vector. 

This approach to capture discourse structure is however quite limited both in general and especially when you consider its application to extractive summarization. First, their hierarchical method has a large number of parameters and it is therefore  slow to train and likely prone to overfitting\footnote{To address this, they only process the first 2000 words of each document, by setting a hard threshold in their implementation, and therefore loosing information.}. Secondly, it  does not take the global context of the whole document into account, which may arguably be critical in extractive methods, when deciding on the salience of a sentence (or even a word). The extractive summarizer we present in this paper tries to address these limitations by adopting the  parameter lean LSTM-minus method, and by explicitly modeling the global context.
\subsection{LSTM-Minus}
The LSTM-Minus method is first proposed in \cite{lstm-minus_propose} as a novel way to learn sentence segment embeddings for graph-based dependency parsing, i.e. estimating the most likely dependency tree given an input sentence. For each dependency pair, they divide a sentence into three segments (prefix, infix and suffix), and LSTM-Minus is used to represent each segment. They apply a single LSTM to the whole sentence and  use the difference between two hidden states $h_j-h_i$ to represent the segment from word  $w_i$  to word $w_j$. This enables  their model to learn segment embeddings from information both outside and inside the segments and thus enhancing their model ability to access to sentence-level information. The intuition behind the method is that each hidden vector $h_t$ can capture useful information before and including the word $v_t$. 

Shortly after, \newcite{lstm-minus_constituency} use the same method on the task of constituency parsing, as the representation of a sentence span, extending the original uni-directional LSTM-Minus to the bi-directional case. 
More recently, inspired by the success of LSTM-Minus in both dependency and constituency parsing, \newcite{lstm-minus_discourse} extend the technique to discourse parsing. They propose a two-stage model consisting of an intra-sentential parser and a multi-sentential parser, learning contextually informed representations of
constituents with LSTM-Minus, at the sentence and document level, respectively. 

Similarly, in this paper, when deciding if a sentence should be included in the summary, the local context of that sentence is captured by applying LSTM-Minus at the document level, to represent the sub-sequence of sentences of the document (i.e., the topic/section) the target sentence belongs to.

\section{Our Model}
\begin{figure*}[t!]
    \centering
    \includegraphics[width=\linewidth]{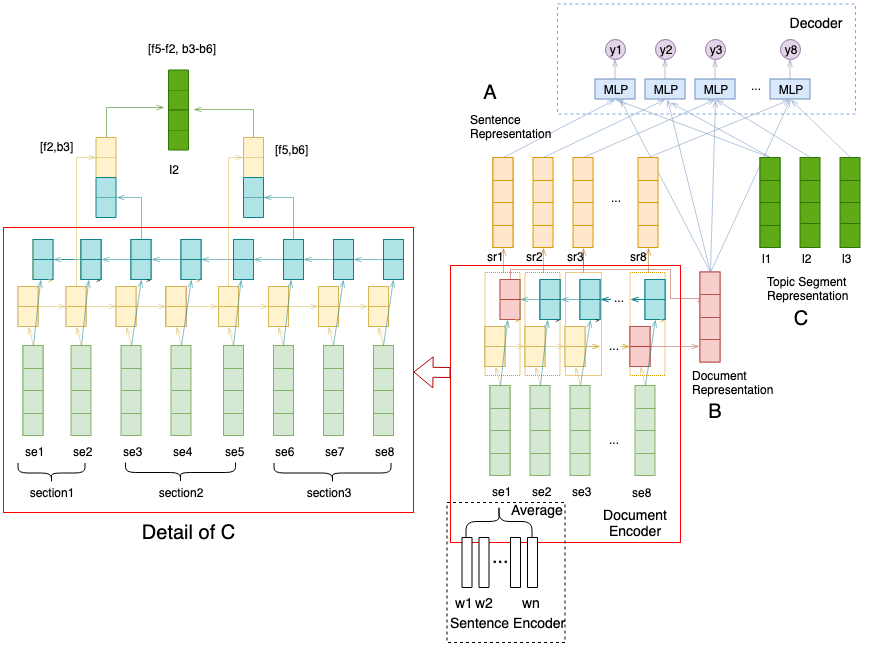}
    \caption{The structure of our model, $se_i,sr_i$ represent the sentence embedding and sentence representation of sentence $i$, respectively. The binary decision of whether the sentence should be included in the summary is based on the sentence itself (A), the whole document (B) and the current topic (C). The document representation is simply the concatenation of the last hidden states of the forward and backward RNNs, while the topic segment representation is computed by applying LSTM-Minus, as shown in detail in the left panel (Detail of C).}
    \label{fig:model}
\end{figure*}
In this work, we propose an extractive model for long documents, incorporating local and global context information, motivated by natural topic-oriented structure of human-written long documents. The architecture of our model is shown in Figure \ref{fig:model}, each sentence is visited sequentially in the original document order, and a corresponding confidence score is computed expressing whether the sentence should be included in the extractive summary.
Our model comprises three components: the sentence encoder, the document encoder and the sentence classifier.
\subsection{Sentence Encoder}
The goal of the sentence encoder is mapping sequences of word embeddings to a fixed length vector (See bottom center of Figure \ref{fig:model}). There are several common 
methods to embed sentences. For  extractive summarization, RNN were used in \cite{summarunner}, CNN in \cite{cheng&lapata}, and Average Word Embedding in \cite{EMNLP2018}. \newcite{EMNLP2018} experiment with all the three methods and conclude that Word Embedding Averaging is as good or better than either RNNs or CNNs for sentence embedding across different domains and summarizer architectures. Thus, we use the Average Word Embedding as our sentence encoder, by which a sentence embedding is simply the average of its word embeddings, i.e. 
\begin{eqnarray*}
se=\frac{1}{n}\sum_{w_0}^{w_n}emb(w_i), se\in \mathbb{R}^{d_{emb}}.
\end{eqnarray*}

Besides, we also tried the popular pre-trained BERT sentence embedding \cite{bert}, but initial results were rather poor. So we do not pursue this possibility any further. 



\subsection{Document Encoder}
At the document level, a bi-directional recurrent neural network \cite{bi-rnn} is often used to encode all the sentences sequentially forward and backward, with such model achieving remarkable success in machine translation \cite{bi-rnn_mt}. As units, we selected gated recurrent units (GRU) \cite{GRU_propose}, in light of favorable results shown in \cite{GRU_why}.  The GRU is represented 
with the  standard  reset, update, and new gates.

The output of the bi-directional GRU for each sentence $t$ comprises two hidden states, $h^f_t \in \mathbb{R}^{d_{hid}},h^b_t \in \mathbb{R}^{d_{hid}}$ as forward and backward hidden state, respectively.\\
\textbf{A. Sentence representation} As shown in Figure \ref{fig:model}(A), for each sentence $t$, the sentence representation is the concatenation of both backward and forward hidden state of that sentence.  $$sr_t = (h^f_t:h^b_t),   sr_t\in\mathbb{R}^{d_{hid}*2}$$ In this way, the sentence representation not only represents the current sentence, but also partially covers contextual information both before and after this sentence. \\
\textbf{B. Document representation}
The document representation provides global information on the whole document. It is computed as the concatenation of the final state of the forward and backward GRU, labeled as B in Figure \ref{fig:model}. \cite{selection_emnlp2018}
\begin{eqnarray*}
d = (h^f_{n}:h^b_0), d\in\mathbb{R}^{d_{hid}*2}
\end{eqnarray*}
\textbf{C. Topic segment representation}
To capture the local context of each sentence, namely the information of the topic segment that sentence falls into, we apply the LSTM-Minus method\footnote{In the original paper, LSTMs were used as recurrent unit. Although we use GRUs here, for consistency with previous work, we still call the method LSTM-Minus}, a method for learning embeddings of text spans. LSTM-Minus is shown in detail in Figure 1 (left panel C), each topic segment is represented as the subtraction between the hidden states of the start and the end of that topic. As illustrated in Figure \ref{fig:model}, the representation for section 2 of the sample document (containing three sections and eight sentences overall) can be computed as $[f_5-f_2,b_3-b_6]$, where $f_5, f_2$ are the forward hidden states of sentence $5$ and $2$, respectively, while $b_3, b_6$ are the backward hidden states of sentence $3$ and $6$, respectively. In general, the topic segment representation  $l_t$ for segment $t$ is computed as:
\begin{eqnarray*}
f_t&=& h^f_{end_t}- h^f_{start_t-1}, f_t \in\mathbb{R}^{d_{hid}}\\
b_t&=& h^b_{start_t}- h^b_{end_t+1}, b_t \in\mathbb{R}^{d_{hid}}\\
l_t&=& (f_t:b_t), l_t \in\mathbb{R}^{d_{hid}*2}
\end{eqnarray*}
where $start_t, end_t$ is the index of the beginning and the end of topic $t$, $f_t$ and $b_t$ denote the topic segment representation of forward and backward, respectively. The final representation of topic $t$ is the concatenation of forward and backward representation $l_t$. To obtain $f_i$ and $b_i$, we utilize subtraction between GRU hidden vectors of $start_t$ and $end_t$, and we pad the hidden states with zero vectors both in the beginning and the end, to ensure the index can not be out of bound. The intuition behind this process is that the GRUs can keep previous useful information in their
memory cell by exploiting reset, update, and new gates to decide how to utilize and update the memory of previous information. In this way, we can represent the contextual information within each topic segment for all the sentences in that segment.
\subsection{Decoder}
Once we have obtained a representation for the sentence, for its topic segment (i.e., local context) and for the document (i.e., global context), these three factors are combined to make a final prediction $p_i$ on whether the sentence should be included in the summary. We consider two ways in which these three factors can be combined.\\
\textbf{Concatenation} We can simply concatenate the vectors of these three factors as, 
$$input_i = (d:l_t:sr_i), input_i \in\mathbb{R}^{d_{hid}*6}$$
where sentence $i$ is part of the topic $t$, and $input_i$ is the representation of sentence $i$ with topic segment information and global context information.\\
\textbf{Attentive context} As local context and global context are all contextual information of the given sentence, we use an attention mechanism to decide the weight of each context vector, represented as
\begin{eqnarray*}
score^d_i &=& v^Ttanh(W_a (d:sr_i))\\
score^l_i &=& v^Ttanh(W_a (l_t:sr_i))\\
weight^d_i &=& \frac{score^d_i}{score^d_i+score^l_i}\\
weight^l_i &=& \frac{score^l_i}{score^d_i+score^l_i}\\
context_i &=& weight^d_i*d+weight^l_i*l_t\\
input_i &=& (sr_i:context_i), input_i \in \mathbb{R}^{d_{hid}*4}
\end{eqnarray*}
where the $context_i$ is the weighted context vector of each sentence $i$, and assume sentence $i$ is in topic $t$.

Then there is a final multi-layer perceptron(MLP) followed with a sigmoid activation function indicating the confidence score for selecting each sentence: 
\begin{eqnarray*}
h_i &=& Dropout(ReLU(W_{mlp} input_i+b_{mlp}))\\
p_i &=& \sigma(W_h h_i +b_h)
\end{eqnarray*}

\section{Experiments}
To validate our method, we set up experiments on the two scientific paper datasets (arXiv and PubMed). With ROUGE and METEOR scores as automatic evaluation metrics, we compare with previous works, both abstractive and extractive.
\subsection{Training}
The weighted negative log-likelihood is minimized, where the weight is computed as $w_{pos} = \frac{\# negative}{\# postive}$, to solve the problem of highly imbalanced data (typical in extractive summarization).
\begin{eqnarray*}
    \mathcal{L} &=& -\sum_{d=1}^{N}\sum_{i=1}^{N_d}(w_{pos} * y_i \log p(y_i|\mathbf{W},b) \\
    &+& (1-y_i)\log p(y_i|\mathbf{W},b))
\end{eqnarray*}
where $y_i$ represent the ground-truth label of sentence $i$, with $y_i=1$ meaning  sentence $i$ is in the gold-standard extract summary. 


\subsection{Extractive Label Generation}
In the Pubmed and arXiv datasets, the extractive summaries are missing. So we follow the work of \cite{EMNLP2018} on extractive summary labeling, constructing gold label sequences by greedily optimizing ROUGE-1 on the gold-standard abstracts, which are available for each article. \footnote{For this, we use a popular python implementation of the ROUGE score to build the oracle. Code can be found here, \url{https://pypi.org/project/py-rouge/}} The algorithm is shown in Appendix A.
\subsection{Implementation Details}
We train our model using the Adam optimizer \cite{adam} with learning rate $0.0001$ and a drop out rate of 0.3. We use a mini-batch with a batch size of 32 documents, and the size of the GRU hidden states is 300. For word embeddings, we use GloVe \cite{glove} with dimension 300, pre-trained on the Wikipedia and Gigaword. The vocabulary size of our model is 50000. All the above parameters were set based on \cite{EMNLP2018} without any fine-tuning. Again following \cite{EMNLP2018}, we train each model for 50 epochs, and the best model is selected with early stopping on the validation set according to Rouge-2 F-score.
\subsection{Models for Comparison}
We perform a systematic comparison with previous work in extractive summarization. For completeness, we also compare with recent neural abstractive approaches. In all the experiments, we use the same train/val/test splitting. 
\begin{itemize}
\setlength\itemsep{-0.3em}
\item Traditional extractive summarization models: SumBasic \cite{sumbasic}, LSA \cite{lsa}, and LexRank \cite{lexrank}
\item Neural abstractive summarization models: Attn-Seq2Seq \cite{RNN_beyond}, Pntr-Gen-Seq2Seq \cite{get_to_the_point} and Discourse-aware \cite{discourse_long_document}
\item Neural extractive summarization models: 
    Cheng\&Lapata \cite{cheng&lapata} and SummaRuNNer \cite{summarunner}. Based on 
    \cite{EMNLP2018}, we use the Average Word Encoder as sentence encoder for both models, instead of the CNN and RNN sentence encoders that were originally used in the two systems, respectively.
    \footnote{Aiming for a fair and reproducible comparison, we re-implemented the models by borrowing the extractor classes from \cite{EMNLP2018}, the source code can be found \url{https://github.com/kedz/nnsum/tree/emnlp18-release}}
\item Baseline: Similar to our model, but without local context and global context, i.e. the input to MLP is the sentence representation only.
\item Lead: Given a length limit of $k$ words for the summary, Lead will return the first $k$ words of the source document. 
\item Oracle: uses the Gold Standard extractive labels, generated based on ROUGE (Sec. 4.2).
\end{itemize}
\subsection{Results and Analysis}

\begin{table}[t!]
    \centering
    \resizebox{\linewidth}{!}{
    \begin{tabular}{|ccccc|}
    \hline
      \it{Model}   & \it{ROUGE-1} & \it{ROUGE-2} & \it{ROUGE-L} &\it{METEOR}\\
      \hline
      SumBasic*   & 29.47& 6.95 &26.30&-\\
      \hline
      LSA* & 29.91 &7.42& 25.67&-\\
      \hline
      LexRank* & 33.85& 10.73&28.99&-\\
      \hhline{=====}
      Attn-Seq2Seq* & 29.30 & 6.00 & 25.56&-\\
      \hline 
      Pntr-Gen-Seq2Seq* & 32.06 & 9.04& 25.16&-\\
      \hline
      Discourse-aware*  & 35.80 & 11.05 &\textbf{31.80}&-\\
      \hhline{=====}
      Baseline &42.91 &16.65 &28.53&21.35\\
      \hline
      Cheng \& Lapata &42.24 & 15.97&27.88&20.97\\
      \hline
      SummaRuNNer & 42.81&16.52 &28.23&21.35\\
      \hline
      Ours-attentive context & \textbf{43.58}&\textbf{17.37}&\textbf{29.30}&\textbf{21.71}\\
      \hline
      Ours-concat & \textbf{43.62}&\textbf{17.36} &\textbf{29.14}&\textbf{21.78}\\
      \hhline{=====}
      Lead &33.66 & 8.94&22.19&16.45\\
      \hline
      Oracle &53.88 & 23.05& 34.90&24.11\\
      \hline
      
    \end{tabular}}
    \caption{Results on the arXiv dataset. For models with an $*$, we report results from \cite{discourse_long_document}. Models are traditional extractive in the first block, neural abstractive in the second block, while neural extractive in the third block. The Oracle (last row) corresponds to using the ground truth labels, obtained (for training) by the greedy algorithm, see Section 4.2. Results that are not significantly distinguished from the best systems are bold.}
    \label{tab:result-arXiv}
\end{table}

\begin{table}[t!]
    \centering
    \resizebox{\linewidth}{!}{
    \begin{tabular}{|ccccc|}
    \hline
    
      \it{Model}   & \it{ROUGE-1} & \it{ROUGE-2} & \it{ROUGE-L} &\it{METEOR}\\
      \hline

      SumBasic*   & 37.15 &11.36 &33.43& -\\
      \hline
      LSA* &33.89 &9.93&29.70& -\\
      \hline
      LexRank* & 39.19 & 13.89 &34.59& -\\
      \hhline{=====}
      Attn-Seq2Seq* &31.55&8.52&27.38&-\\
      \hline 
      Pntr-Gen-Seq2Seq* & 35.86 &10.22&29.69&-\\
      \hline
      Discourse-aware*  & 38.93&15.37 & \textbf{35.21}&-\\
      \hhline{=====}
      Baseline & 44.29&19.17 &30.89&20.56\\
      \hline
      Cheng \& Lapata &43.89 & 18.53&30.17&20.34\\
      \hline
      SummaRuNNer & 43.89 &18.78  & 30.36&20.42\\
      \hline
      Ours-attentive context &\textbf{44.81} &\textbf{19.74} &\textbf{31.48}&\textbf{20.83}\\
      \hline
      Ours-concat & \textbf{44.85}&\textbf{19.70} &\textbf{31.43}&\textbf{20.83}\\
      \hhline{=====}
      Lead & 35.63 &12.28&25.17&16.19\\
      \hline
      Oracle &55.05 &27.48 &38.66&23.60\\
      \hline
      
    \end{tabular}}
    \caption{Results on the Pubmed dataset. 
    See caption of Table \ref{tab:result-arXiv} above for details on compared models and notation. 
    }
    \label{tab:result-pubmed}
\end{table} 
For evaluation, we follow the same procedure as in \cite{EMNLP2018}. Summaries are generated by selecting the top ranked sentences by model probability $p(y_i|\mathbb{W},b)$, until the length limit is met or exceeded. Based on the average length of abstracts in these two datasets, we set the length limit to 200 words. We use ROUGE scores\footnote{We use a modified version of rouge\_papier, a python wrapper of ROUGE-1.5.5, \url{https://github.com/kedz/rouge_papier}. The command line is 'Perl ROUGE-1.5.5 -e data -a -n 2 -r 1000 -f A -z SPL  config\_file'} \cite{ROUGE} and METEOR scores\footnote{We use default setting of METEOR.} \cite{meteor} between the model results and ground-truth abstractive summaries as evaluation metric. The unigram and bigram overlap (ROUGE-1,2) are intended to measure the informativeness, while longest common subsequence (ROUGE-L) captures fluency to some extent \cite{cheng&lapata}. METEOR was originally proposed to evaluate translation systems by measuring the alignment between the system output and reference translations. As such, it can also be used as an automatic evaluation metric for summarization \cite{EMNLP2018}.
\begin{figure*}[htbp!]
    \centering
    \includegraphics[width=\linewidth]{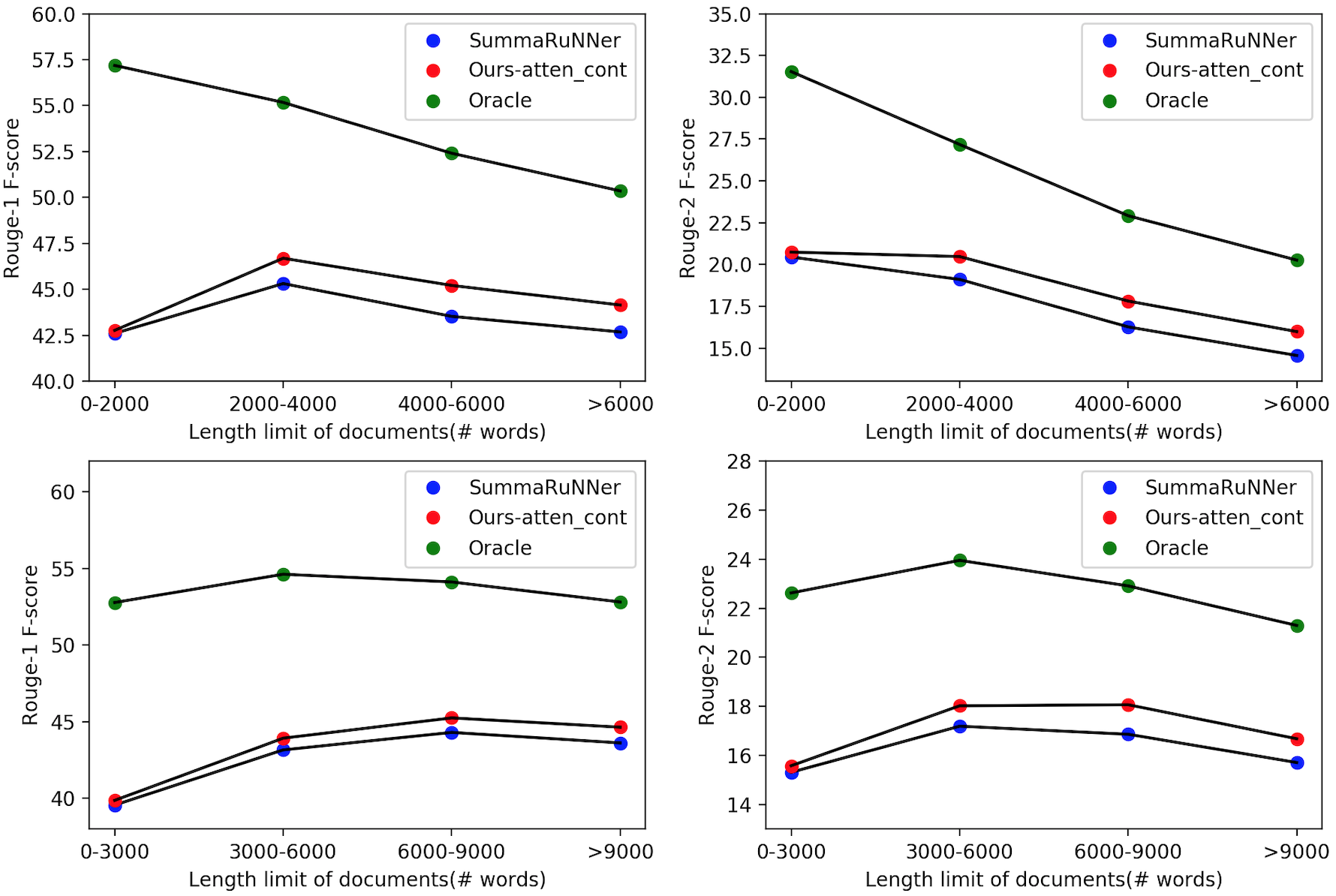}
    \caption{A Comparison between our model, SummaRuNNer and Oracle when applied to documents with increasing length, left-up: ROUGE-1 on Pubmed dataset, right-up: ROUGE-2 on Pubmed dataset, left-down: ROUGE-1 on arXiv dataset, right-down: ROUGE-2 on arXiv dataset}
    \label{fig:long docs}
\end{figure*} \squeezeup

The performance of all models on  arXiv  and Pubmed is shown in Table \ref{tab:result-arXiv} and Table \ref{tab:result-pubmed}, respectively. Follow the work \cite{EMNLP2018}, we use the approximate randomization as the statistical significance test method \cite{statsig} with a  Bonferroni correction for multiple comparisons, at the confidence level 0.01 ($p<0.01$).  
As we can see in these tables, on both datasets, the neural extractive models outperforms  the traditional extractive models on informativeness (ROUGE-1,2) by a wide margin, but results are mixed on ROUGE-L. Presumably, this is due to the neural training process, which relies on a goal standard based on ROUGE-1. Exploring other training schemes and/or a combination of traditional and neural approaches is left as future work. Similarly, the neural extractive models also dominate the neural abstractive models on ROUGE-1,2, but these abstractive models tend to have the highest ROUGE-L scores, possibly because they are trained directly  on gold standard abstract summaries.




Compared with other neural extractive models, our models (both with attentive context and concatenation decoder) have better performances on all three ROUGE scores, as well as METEOR. In particular, the improvements over the Baseline model show that a combination of local and global contextual information does help to identify the most important sentences (more on this in the next section). Interestingly, just the Baseline model already achieves a slightly better performance than  previous works; possibly because the auto-regressive approach used in those models is  even more detrimental for long documents.


Figure \ref{fig:long docs} shows the most important result of our analysis: the benefits of our method, explicitly designed 
to deal with longer documents, do actually become stronger as we apply it to longer documents. As it can be seen in Figure \ref{fig:long docs}, the performance gain of our model with respect to current state-of-the-art extractive summarizer is more pronounced for documents with $>= 3000$ words in both datasets. 

Finally, the result of Lead (Table \ref{tab:result-arXiv}, \ref{tab:result-pubmed}) shows that scientific papers have less position bias than news; i.e., the first  sentences of these papers are not a good choice to form an extractive summary. 

As a teaser for the potential and challenges that still face our approach, its output (i.e., the extracted sentences) {\it when applied to this paper} is colored in red and the order in which the sentences are extracted is marked with the Roman numbering. If we set the summary length limit to the length of our abstract, the first five sentences in the conclusions section are extracted. If we increase the length to 200 words, two more sentences are extracted, which do seem to provide useful complementary information. Not surprisingly, some redundancy is present, as dealing explicitly with redundancy is not a goal of our current proposal and left as future work.
\subsection{Ablation Study}

\begin{table}[htb!]
    \centering
    \resizebox{\linewidth}{!}{
    \begin{tabular}{|llll|}
    \hline
    
      \it{Model}   & \it{ROUGE-1(+l/+g)} & \it{ROUGE-2(+l/+g)} & \it{ROUGE-L(+l/+g)} \\
      \hhline{====}
      BSL & 44.29 (na/na)&19.17 (na/na) &30.89 (na/na)\\
      \hline
      BSL+l & \textbf{44.85} (+.56/na) & \textbf{19.77} (+.6/na) & \textbf{31.51} (+.62/na)\\
      \hline
      BSL+g & 44.06 (na/-.23) &18.83 (na/-.34) & 30.53 (na/-.36)\\
      \hline
      BSL+l+g & \textbf{44.81} (+.75/-.04) &\textbf{19.74} (+.91/-.03) &\textbf{31.48} (+.95/-.03)\\
      \hhline{====}
    BSL &43.85 (na/na) & 15.94 (na/na) & 28.13 (na/na)\\
      \hline
      BSL+l & \textbf{44.65} (+.8/na) & \textbf{16.75} (+.81/na)& \textbf{28.99} (+.85/na)\\
      \hline
      BSL+g &43.70 (na/-.15) & 15.74 (na/-.2) & 27.67 (na/-.46)\\
      \hline
      BSL+l+g & \textbf{44.64} (+.94/-.01)&\textbf{16.69} (+.95/-.06)&\textbf{28.96} (+1.29/-.03)\\
      \hline

    \end{tabular}}
    \caption[Ablation study on Pubmed]{Ablation study on the Pubmed dataset, with all the documents(up) and a subset of long documents (down, $>6000$ words). BSL is the model with sentence representation only, BSL+l is the model with sentence  and local topic information, BSL+g is the model with sentence and global document information, and the last one is the full model with attentive\_context decoder. The numbers in parenthesis represent the improvements with the additional local/global context, respectively. Results that are not significantly distinguished from the best systems are bold.}
    \label{tab:ablation-Pubmed}
\end{table}

\begin{table}[htb!]
    \centering
    \resizebox{\linewidth}{!}{
    \begin{tabular}{|llll|}
    \hline
    
      \it{Model}   & \it{ROUGE-1(+l/+g)} & \it{ROUGE-2(+l/+g)} & \it{ROUGE-L(+l/+g)} \\
      \hhline{====}
      BSL &42.91 (na/na) &16.65 (na/na) &28.53 (na/na)\\
      \hline
      BSL+l &\textbf{43.57} (+.66/na) & \textbf{17.35} (+.7/na) & \textbf{29.29} (+.76/na)\\
      \hline
      BSL+g &42.90 (na/-.01) &16.58 (na/-.07) &28.36 (na/-.17)\\
      \hline
      BSL+l+g & \textbf{43.58} (+.68/+.01)&\textbf{17.37} (+.79/+.02)&\textbf{29.30} (+.94/+.01)\\

      \hhline{====}
      BSL & 42.95 (na/na)&14.85 (na/na)&28.66 (na/na)\\
      \hline
      BSL+l & \textbf{44.01} (+1.06/na) & \textbf{15.95} (+1.1/na) & \textbf{29.68} (+1.02/na)\\
      \hline
      BSL+g & 43.05 (na/+.1)&14.91 (na/+.06)&28.57 (na/-.09)\\
      \hline
      BSL+l+g &\textbf{44.17} (+1.12/+.16) & \textbf{16.01} (+1.1/+.06) & \textbf{29.72} (+1.15/+.04)\\
      \hline
      
    \end{tabular}}
    \caption[Ablation study on Pubmed]{Ablation study on arXiv dataset, with all documents (up) and a subset of long document(down, $>9000$ words).  Results that are not significantly different from the best systems are in bold.}
    \label{tab:ablation-arXiv}
\end{table}

In order to  assess  the relative  contributions  of  the  global  and  local models to the performance of our approach, we ran an ablation study. This was done for each dataset both with the whole test set, as well as with a subset of long documents. The results for Pubmed and arXiv are shown in  
 Table \ref{tab:ablation-Pubmed} and Table \ref{tab:ablation-arXiv}, respectively. For statistical significance, as it was done for the general results in Section 4.5, we use the approximate randomization method \cite{statsig} with the Bonferroni correction at ($p<0.01$). 

From these tables, we can see that on both datasets the performance significantly improves when local topic information (i.e. local context) is added. And the improvement is even greater when we only consider long documents. Rather surprisingly,  this is not the case for the global context. Adding a representation of the whole document (i.e. global context) never significantly improves performance. In essence, it seems that all the  benefits of our model come exclusively from modeling  the  local  context,  even  for  the  longest documents. Further investigation of this finding is left as future work.



\section{Conclusions and Future Work}
\renewcommand{\thefootnote}{\Roman{footnote}}
\textcolor{purple}{In this paper, we propose a novel extractive summarization model especially designed for long documents, by incorporating the local context within each topic, along with the global context of the whole document.}\footnotemark[2]  \textcolor{purple}{Our approach integrates recent findings on neural extractive summarization in a parameter lean and modular architecture.}\footnotemark[3]
\textcolor{purple}{We evaluate our model and compare with previous works in both extractive and abstractive summarization on two large scientific paper datasets,
which contain documents that are much longer than in previously used corpora.}\footnotemark[4] \textcolor{purple}{Our model not only achieves state-of-the-art on these two datasets, but in an additional experiment, in which we consider documents with increasing length, it becomes more competitive for longer documents.}\footnotemark[5]
\textcolor{purple}{We also ran an ablation study to assess the relative contribution of the global and local components of our approach. }\footnotemark[1]
Rather surprisingly, it appears that the benefits of our model come only from modeling the local context.

For future work, we initially intend to investigate neural methods to deal with redundancy.
Then, it could be beneficial to integrate explicit features, like sentence position and salience, into our neural approach. More generally, we plan to combine of traditional and neural models, as suggested by our results. Furthermore, we would like to explore more sophistical structure of documents, like discourse tree, instead of rough topic segments. As for evaluation, we would like to elicit human judgments, for instance by inviting authors to rate the outputs from different systems, when applied to their own papers.
More long term, we will study how extractive/abstractive
techniques can be integrated; for instance, the output of an extractive system could be fed into an abstractive
one, training the two jointly.


\section*{Acknowledgments}
This research was supported by the Language \& Speech Innovation Lab of Cloud BU, Huawei Technologies Co., Ltd.
\bibliography{emnlp-ijcnlp-2019}
\bibliographystyle{acl_natbib}

\appendix
\section{Extractive Label Generation}
The algorithm \ref{extractive_label_gen} is used to generate the extractive labels based on the human-made abstractive summaries, i.e. abstracts of scientific papers.
\begin{algorithm*}
\caption{Extractive label generation}\label{extractive_label_gen}
\begin{algorithmic}
\Function{LabelGeneration}{Reference,sentences,lengthLimit}
\State $hyp$ = ''
\State $wc$ = 0
\State	$picked$ = []
\State	$highest\_r1 = 0$
\State $sid = -1$
\While{$wc\leq lengthLimit$}
\For{$i$ in range(len($sentences$))}
\State $score = ROUGE(hyp+sentences[i],ref)$
\If{$score > highest\_r1$}
\State $highest\_r1 = score$
\State $sid = i$
\EndIf
\EndFor
\If{$sid$!=$-1$}
\State $picked$.append($sid$)
\State $hyp$ = $hyp$+ $sentences$[$sid$]
\State $wc$ += NumberOfWords($sentences$[$sid$])
\Else 
\State break
\EndIf
\EndWhile
\State \Return  $picked$
\EndFunction
\end{algorithmic}
\end{algorithm*}
\end{document}